\pdfoutput=1

\documentclass[11pt]{article}

\usepackage{times}
\usepackage{latexsym}
\usepackage{graphicx}
\usepackage{sidecap}
\usepackage[small]{caption}
\usepackage{subcaption}
\usepackage{amsmath}
\allowdisplaybreaks
\usepackage{amsthm}

\usepackage{thmtools}
\usepackage{thm-restate}

\usepackage{multicol}

\usepackage{pifont}
\usepackage{xspace}

\usepackage[round]{natbib}

\usepackage{appendix}

\newcommand{\rightcomment}[1]{\(\triangleright\) {\small \it #1}}
\newcommand{\eqcomment}[1]{\addtocounter{equation}{1}\tag*{\rightcomment{#1}\quad(\theequation)}}
\usepackage{suffix}
\WithSuffix\newcommand\eqcomment*[1]{\tag*{\rightcomment{#1}}}

\usepackage[hidelinks]{hyperref}

\usepackage{EMNLP2022}

\usepackage{xcolor}

\newcommand{\cutforspace}[1]{}

\usepackage[T1]{fontenc}
\usepackage[utf8]{inputenc}
\usepackage{hyperref}       %
\usepackage{url}            %
\usepackage{booktabs}       %
\usepackage{amsfonts}       %
\usepackage{nicefrac}       %
\usepackage{microtype}      %

\usepackage{microtype}

\usepackage{inconsolata}

\usepackage{algorithm}
\usepackage[noend]{algpseudocode}

\renewcommand\algorithmicthen{:}
\algnewcommand{\IfThen}[2]{\State \algorithmicif\ #1\ \algorithmicthen\ #2}
\algnewcommand{\IfThenElse}[3]{\State \algorithmicif\ #1\ \algorithmicthen\ #2\ \algorithmicelse\ #3}
\algrenewcommand{\algorithmiccomment}[1]{\hfill \rightcomment{#1}}
\algnewcommand{\LineComment}[1]{\State \rightcomment{#1}}
\algnewcommand{\LinesComment}[1]{\State \rightcomment{\parbox[t]{\linewidth-\leftmargin-\widthof{\(\triangleright\) }}{#1}}\smallskip}
\algnewcommand\algorithmicinput{{\bfseries Input:}}
\algnewcommand\INPUT{\item[\algorithmicinput]}
\algnewcommand\algorithmicoutput{{\bfseries Output:}}
\algnewcommand\OUTPUT{\item[\algorithmicoutput]}
\makeatletter
\newcounter{algorithmicH}
\let\oldalgorithmic\algorithmic
\renewcommand{\algorithmic}{%
  \stepcounter{algorithmicH}
  \oldalgorithmic}
\renewcommand{\theHALG@line}{ALG@line.\thealgorithmicH.\arabic{ALG@line}}
\makeatother
\makeatletter
\newcommand{\algmargin}{\the\ALG@thistlm}
\makeatother
\algnewcommand{\Statepar}[1]{\State\parbox[t]{\dimexpr\linewidth-\algmargin}{\strut #1\strut}}

\usepackage{xpatch}
\usepackage[compact]{titlesec}
\titlespacing{\section}{0pt}{1ex}{0.5ex}
\titlespacing{\subsection}{0pt}{0.5ex}{0ex}
\titlespacing{\subsubsection}{0pt}{0.5ex}{0ex} 
\renewcommand{\paragraph}[1]{\textbf{#1}}
\usepackage{setspace}
\AtBeginDocument{%
  \addtolength\abovedisplayskip{-0.25\baselineskip}%
  \addtolength\belowdisplayskip{-0.25\baselineskip}%
  \addtolength\abovedisplayshortskip{-0.25\baselineskip}%
  \addtolength\belowdisplayshortskip{-0.25\baselineskip}%
}

\usepackage{amsmath}
\usepackage{amssymb}
\usepackage{mathrsfs}
\usepackage{mathtools, cuted}
\usepackage{tabularx, booktabs}
\newcolumntype{C}{>{\centering\arraybackslash}X}
\newcolumntype{R}{>{\raggedleft\arraybackslash}X}
\newcolumntype{S}{>{\raggedleft\arraybackslash\hsize=.5\hsize}X}
\usepackage{latexsym}
\usepackage{url}
\usepackage{xspace}
\usepackage{bm,array}
\usepackage{amsfonts}
\usepackage{enumitem}
\usepackage{cases}
\usepackage{mathtools}
\usepackage{empheq}
\usepackage{bm}
\usepackage{esvect}
\usepackage[noabbrev,capitalize]{cleveref}
\crefname{equation}{equation}{equations}
\crefname{section}{section}{sections}
\crefname{footnote}{footnote}{footnotes}   
\crefname{line}{line}{lines}   
\crefname{assumption}{assumption}{assumptions}

\usepackage{bbm}

\let\frac=\tfrac  %

\usepackage[compact]{titlesec}  %

\renewcommand{\vec}[1]{{\boldsymbol{\mathbf{#1}}}}   %

\newcommand{\defeq}{\mathrel{\stackrel{\textnormal{\tiny def}}{=}}}

\newcommand{\Real}{\mathbb{R}}

\renewcommand{\th}{\textsuperscript{th}\xspace}

\usepackage{tikz}
\usetikzlibrary{shapes.geometric}

\usetikzlibrary{arrows,decorations.markings}
\usetikzlibrary{arrows}
\newcommand{\blueline}{\begin{tikzpicture} \draw[arrows={-angle 60}, white, thick, rotate=180, opacity=1.0] (0,0.00) -- (0.5,0.00); \draw[arrows={-angle 60}, blue, thick, rotate=180] (0,-0.06) -- (0.5,-0.06); \end{tikzpicture}\xspace}
\newcommand{\redline}{\begin{tikzpicture} \draw[arrows={-angle 60}, white, thick, rotate=180, opacity=1.0] (0,0.00) -- (0.5,0.00); \draw[arrows={-angle 60}, red, thick, rotate=180] (0,-0.06) -- (0.5,-0.06); \end{tikzpicture}\xspace}

\usetikzlibrary{shapes.geometric}
\definecolor{mypurple}{RGB}{153,102,255}
\definecolor{myblue}{RGB}{153,102,255}
\definecolor{mygreen}{RGB}{100,209,150}
\newcommand{\purplestar}{\begin{tikzpicture} \path node[star,star point ratio=2.25,minimum size=6pt, inner sep=0pt,draw=mypurple,solid,fill=mypurple]{}; \end{tikzpicture}\xspace}

\usepackage{verbatim}
\newcommand{\done}[1]{} %

\usepackage{dcolumn}
\newcolumntype{d}[1]{D{/}{/}{#1}}
\newcolumntype{P}[1]{>{\centering\arraybackslash}p{#1}}

\title{Tiny-Attention Adapter:\\
Contexts Are More Important Than the Number of Parameters}

\author{Hongyu Zhao\thanks{\ \ Work done while during internship at TTI-Chicago.} \\
  University of Chicago \\
  \texttt{hongyuz@uchicago.edu} \\\And
  Hao Tan \\
  Adobe Research \\
  \texttt{hatan@adobe.com} \\\And
  Hongyuan Mei \\
  TTI-Chicago \\
  \texttt{hongyuan@ttic.edu} 
  }

\begin{document}
\maketitle
\begin{abstract}
Adapter-tuning is a paradigm that transfers a pretrained language model to downstream tasks by adding and tuning a small number of new parameters. 
Previously proposed adapter architectures are all feed-forward neural networks. 
In this paper, we investigate the effectiveness of using \emph{tiny-attention}---i.e., attention with extremely small per-head dimensionality---as adapters. Our tiny-attention adapter learns to modify the hidden states at each position directly conditioned on the hidden states at all the other positions, which is missed by the previously proposed adapters. 
Moreover, we view its multiple attention heads as a mixture of experts and propose to average their weights during deployment, which further reduces its inference computation cost. 
On the GLUE benchmark, our tiny-attention adapter outperforms the other parameter-efficient transfer learning methods as well as full fine-tuning while only updating 0.05\% of the parameters. 
On the FewGLUE benchmark, its performance is comparable to that of GPT-3 and PET. 
\end{abstract}
\section{Introduction}
\label{sec:intro}
Transferring a large pretrained language model (PLM) is a de facto paradigm of performing downstream tasks in natural language processing (NLP). 
A general approach is adapter-tuning, which means inserting \emph{adapters}---i.e., neural networks with small numbers of parameters---into each pretrained layer and only updating the adapter parameters.
Adapter-tuning is parameter-efficient and enjoys low computation cost since it keeps the PLM frozen. 
But it underperforms full fine-tuning which updates all the parameters of the PLM.
\begin{figure}[!ht]
    \centering
    \includegraphics[width=\columnwidth]{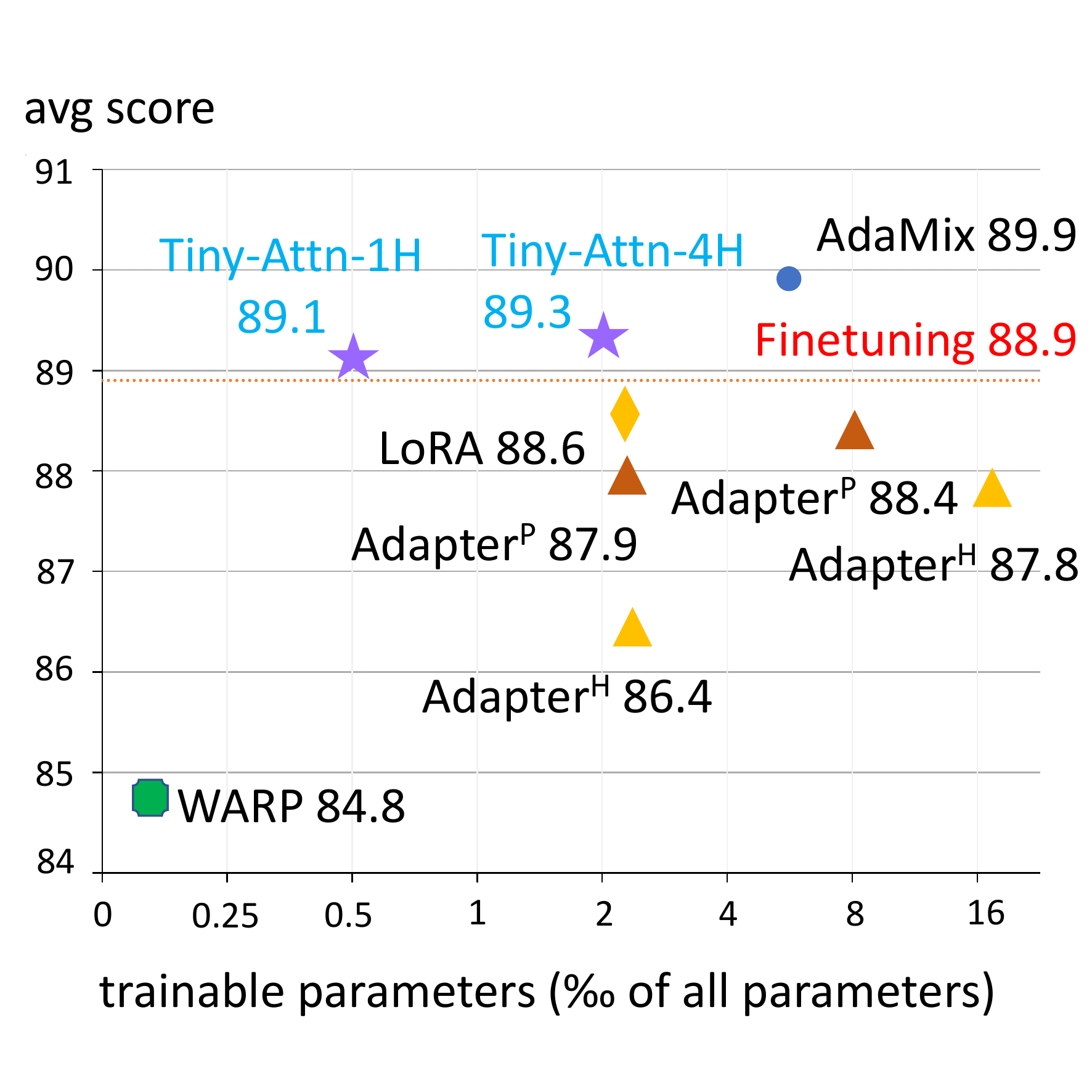}
    \vspace{-32pt}
    \caption{Average performance of different parameter-efficient transfer learning methods on the GLUE benchmark with roberta-large as the PLM. Our method is Tiny-Attn (with \purplestar marker): ``1H'' and ``4H'' mean ``one attention head'' and ``four attention heads (with the parameter-averaging trick in \cref{sec:mix})'' respectively.
    }
    \label{fig:para-count}
\end{figure}
\begin{figure*}[!t]
     \centering
     \begin{subfigure}[b]{0.3\textwidth}
         \centering
         \includegraphics[width=\textwidth]{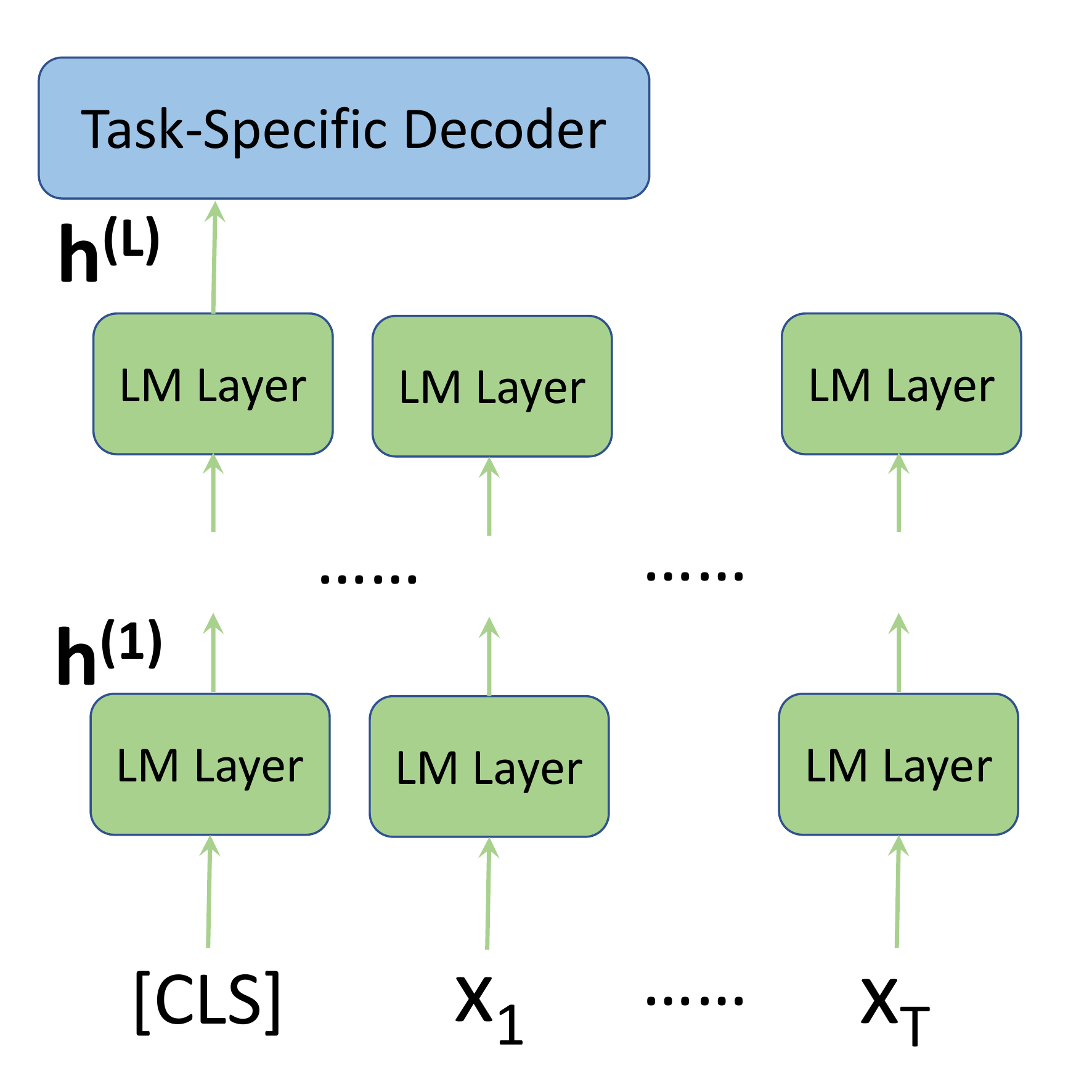}
         \vspace{-20pt}
         \caption{Transferring a pretrained language model for downstream tasks.}
         \label{fig:structure1}
     \end{subfigure}
     \hfill
     \begin{subfigure}[b]{0.3\textwidth}
         \centering
         \includegraphics[width=\textwidth]{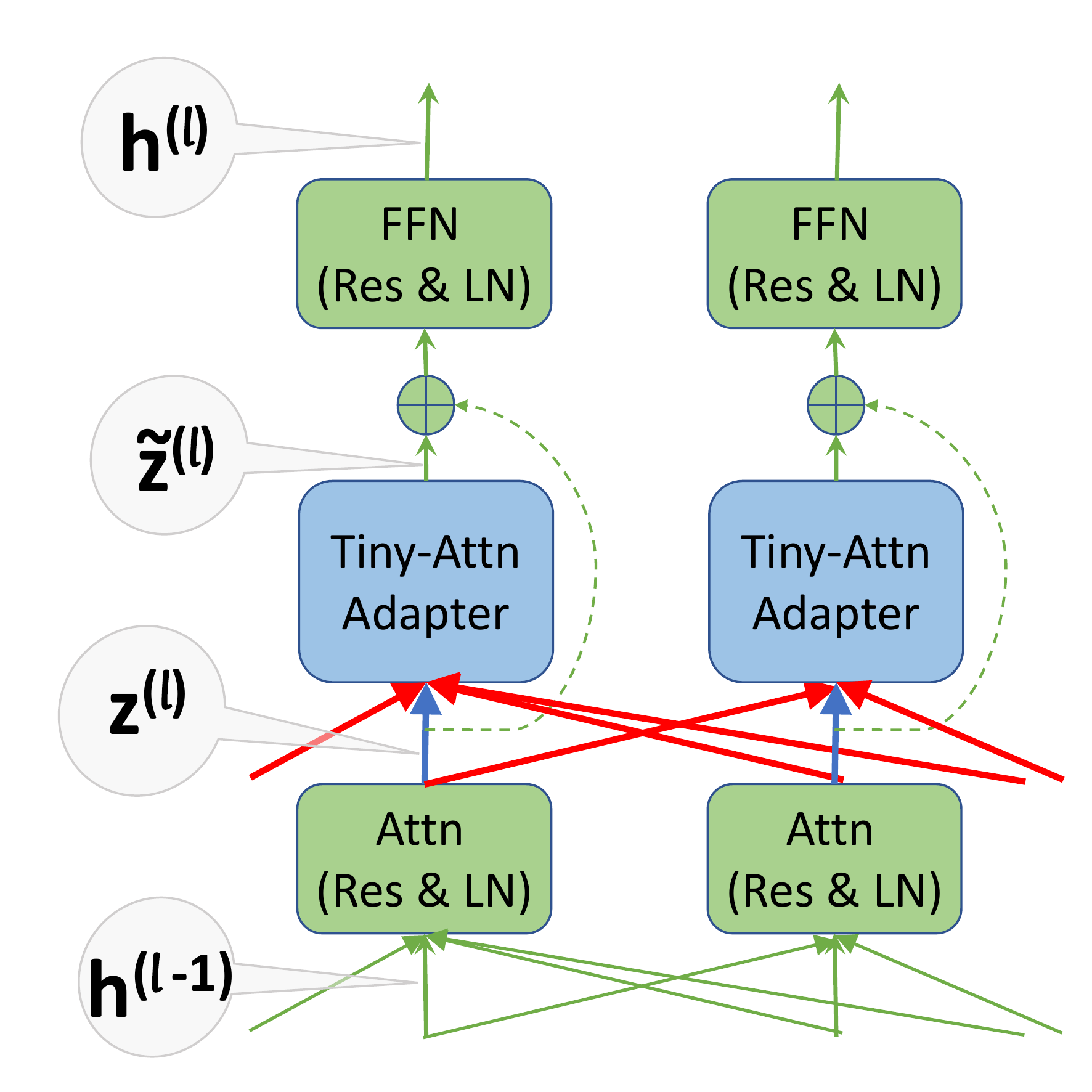}
         \vspace{-20pt}
         \caption{Adapting a language model layer by adding our tiny-attention adapter.  }
         \label{fig:structure2}
     \end{subfigure}
     \hfill
     \begin{subfigure}[b]{0.3\textwidth}
         \centering
         \includegraphics[width=\textwidth]{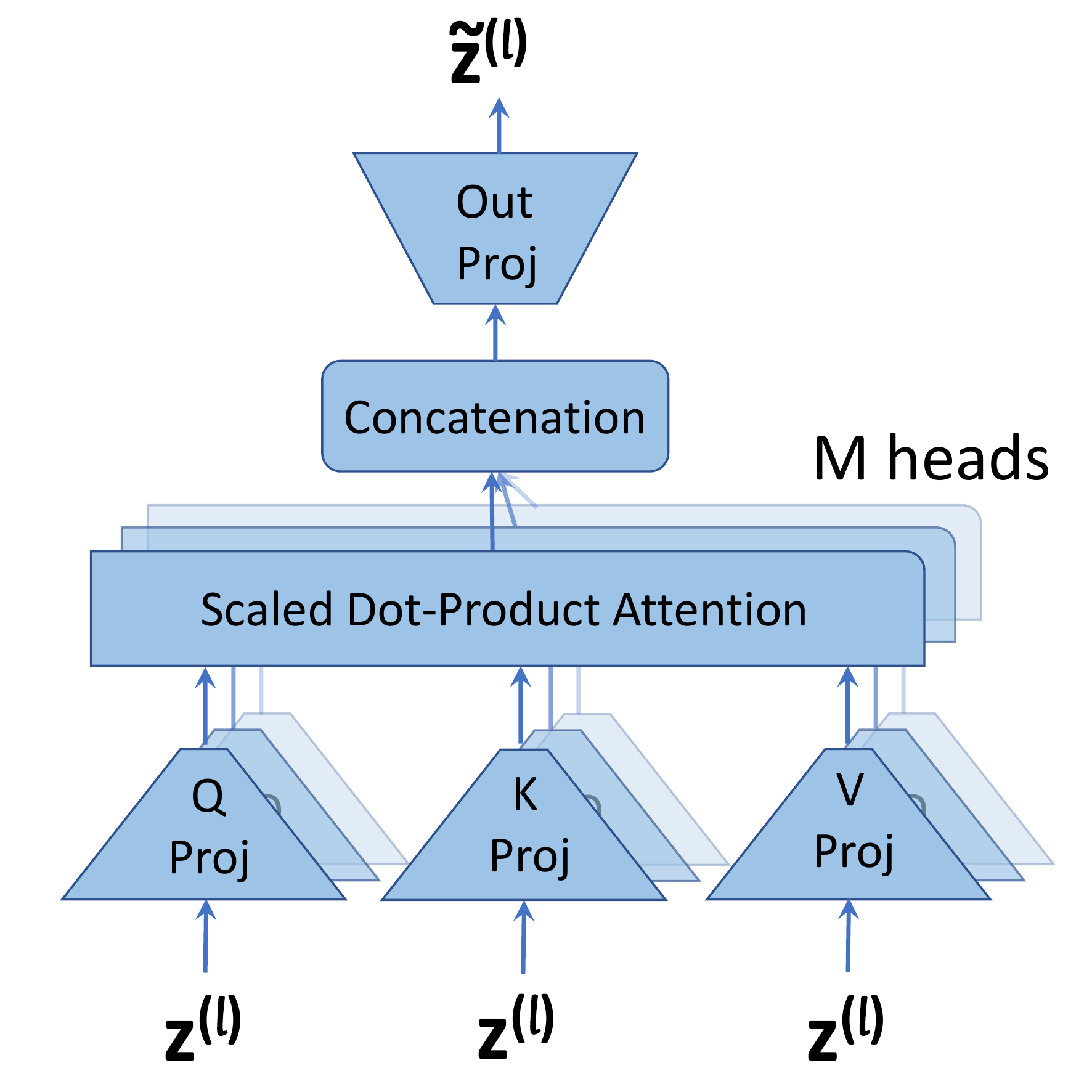}
         \vspace{-20pt}
         \caption{The internal architecture of our tiny-attention adapter. }
         \label{fig:structure3}
     \end{subfigure}
    \vspace{-8pt}
    \caption{The pipeline of applying our tiny-attention adapter to a pretrained language model for downstream tasks.}
    \label{fig:structure}
    \vspace{-8pt}
\end{figure*}

In this paper, we propose a new adapter architecture which outperforms full fine-tuning yet uses even fewer parameters than the previously proposed adapters as well as other parameter-efficient transfer learning methods; see \cref{fig:para-count}. 

Our adapter is a multi-head attention module and its per-head dimensionality is extremely small; thus we call it \emph{tiny-attention}.
The architecture design is inspired by the following intuitions:  
\begin{enumerate}[leftmargin=*,noitemsep]
    \item For each input sequence, each layer of the language model produces an embedding for each token in the sequence; see \cref{fig:structure1}. 
    \item All the parameter-efficient transfer learning methods learn to modify the embeddings towards the direction of performing the given task well; see \citet{he2021towards} for a thorough theoretical and empirical discussion. 
    \item Almost all the previously proposed adapter architectures are feed-forward neural networks. Thus, we suspect that their embedding modifications are not as contextually rich as they should. 
\end{enumerate}
Therefore, we propose to use the attentative structure that allows the embedding modifications of each token to capture more contextual information by directly looking at the embeddings of all the tokens.  
The dimensionality of each attention head need not to be large.
In NLP tasks, contextual information is often demonstrated more important than model sizes: for example, in language modeling, a smaller model with a larger context window usually outperforms a larger model with a smaller context window~\citep{dai-etal-2019-transformer,wu2022memorizing,yang2019xlnet}. 
Additionally, we view the multiple attention heads of our tiny-attention adapter as a mixture of experts and then propose to average their weights during inference. 
This technique further reduces the inference cost. 

We evaluated our tiny-attention adapter on the GLUE~\citep{wang2018glue} and FewGLUE~\citep{schick-schutze-2021-just,superglue} benchmarks. 
On GLUE, it updates only 0.05\% of the parameters---an order of magnitude smaller than all the other methods---yet still outperforms full fine-tuning and nearly all the other parameter-efficient tuning methods. 
On FewGLUE, it is comparable to several strong competing methods including PET~\citep{schick-schutze-2021-just} and GPT-3~\citep{gpt3}.
We also conducted ablation studies to investigate its effectiveness with varying placements and PLMs. 
\section{The Method}\label{sec:method}
\cref{fig:structure1} illustrates how a language model performs a downstream task. 
The language model has $L$ layers. 
Given an input sequence $\vec{x}=x_0x_1\ldots x_T$ where $x_0$ is a special classification (CLS) token, each layer $\ell$ reads the embeddings given by the previous layer $\ell-1$ and produces the layer-$\ell$ embeddings $\vec{h}_{0}^{(\ell)} \vec{h}_{1}^{(\ell)} \ldots \vec{h}_{T}^{(\ell)}$. 
Then a task-specific decoder%
reads the top-layer embedding $\vec{h}_{0}^{(L)}$ of the CLS token and predicts a task-specific output $\hat{y}$ for that sequence. 

Transferring the language model involves updating its \emph{trainable} parameters to minimize a task-specific $\text{loss}(\hat{y},y)$ where $y$ is the ground-truth label for the given $\vec{x}$. 
For adapter-tuning, the trainable parameters only include those of the task-specific decoder and those of the adapters. 
\cref{fig:structure2} shows a language model layer with our tiny-attention adapter placed between its attention module and feed-forward net: during training, only the decoder and adapter parameters (\textcolor{blue}{blue}) are updated while the pretrained parameters (\textcolor{teal}{green}) are all kept frozen.%

\subsection{Tiny-Attention Adapter}\label{sec:tiny}
Our adapter has an attentative structure: as shown in \cref{fig:structure2}, at each position $t$, it takes as input the intermediate embeddings $\vec{z}^{(\ell)}$ from not only the current position (information flow indicated by blue arrows \blueline)
but also all the other positions (information flow shown by red arrows \redline). 
For each token $t$, it produces a task-specific modification $\tilde{\vec{z}}^{(\ell)}_t$. 
Then the modified embeddings ${\vec{z}}^{(\ell)}_t + \tilde{\vec{z}}^{(\ell)}_t$ are fed into the layer $\ell$ feed-forward net for producing $\vec{h}_{t}^{(\ell)}$.

As shown in \cref{fig:structure3}, the internal architecture of our tiny-attention adapter resembles an ordinary multi-head attention mechanism. 
Suppose it has $M$ attention heads. Each attention head $m$ produces a head-specific attention vector $\tilde{\vec{z}}^{(\ell,m)}_t$ and the final attention vector $\tilde{\vec{z}}^{(\ell)}_t$ is obtained by projecting the concatenation of all the $\tilde{\vec{z}}^{(\ell,m)}_t$: 
\begin{subequations}
\begin{align}
    \tilde{\vec{z}}^{(\ell)}_t 
    &\defeq \vec{O}^{(\ell)}[\tilde{\vec{z}}^{(\ell,1)}_t;\ldots;\tilde{\vec{z}}^{(\ell,M)}_t] \label{eq:outproj}\\
    \tilde{\vec{z}}^{(\ell,m)}_t
    &\defeq \text{Attn}_t^{(m)}( {\vec{z}}^{(\ell)}_0, {\vec{z}}^{(\ell)}_1, \ldots, {\vec{z}}^{(\ell)}_T ) \label{eq:attnm}
\end{align}
\end{subequations}
The math details of Attn$_t^{(m)}$ is given in \cref{app:math}.

\paragraph{Why \emph{attention} as adapter?}
Attention allows the task-specific modification $\tilde{\vec{z}}^{(\ell)}_t$ for each token $t$ to aggregate useful information from its full contexts at $t=0,1,\ldots,T$. 
It is analogous to how the attention modules in the pretrained language model learned to construct contextual representations which helped optimize the language modeling objective during pretraining. 
Therefore, when the pretrained attention modules are frozen, 
it seems natural to adopt new trainable attention modules for their desired behavior. 

The key difference between our tiny-attention and an ordinary attention is that our per-head dimension is \emph{tiny}: i.e., $\tilde{\vec{z}}_t^{(\ell,m)} \in \Real^{D}$ and $D$ is very small. 
Throughout our experiments, we set $D=1$. 

\paragraph{Why \emph{tiny}-attention?}
Smaller dimensionality means fewer trainable parameters and less computation cost; thus it is preferred. 
We believe that small dimensionality is sufficient in our setting because of two key observations. 
First, a smaller model often tends to work as well as a larger model if it is allowed to use a larger context; see \cref{sec:intro} for the example of language modeling~\citep{dai-etal-2019-transformer,wu2022memorizing,yang2019xlnet}. 
Second, \citet{hu2021lora} found that feed-forward adapters can achieve competitive results with extremely low-rank (1 or 2) parameter matrices.
Similar to the LoRA method by \citet{hu2021lora}, our tiny-attention adapter essentially performs a low-rank non-linear projection: it first linearly transforms the high-dimensional embeddings $\vec{z}^{(\ell)}$ to low-dimensional query, key, and value vectors; it then linearly transforms the low-dimensional attention vectors---after the non-linear attention operation---to the high-dimensional modification vectors $\tilde{\vec{z}}^{(\ell)}$%
; see \cref{app:math} for technical details.

\subsection{Multiple Heads as a Mixture of Experts}\label{sec:mix}
The multiple attention heads in our tiny-attention can be regarded as a mixture of experts where each head is an expert that specializes in capturing certain kinds of contextual information (e.g., syntax, semantics).
The output projection in \cref{eq:outproj} learns to aggregate the information $\tilde{\vec{z}}_{t}^{(\ell,m)}$ produced by the experts.
Rearranging that equation gives
\begin{align}
    \tilde{\vec{z}}^{(\ell)}_t 
    &\defeq \sum_{m=1}^{M} \vec{O}^{(\ell,m)} \tilde{\vec{z}}^{(\ell,m)}_t \label{eq:outproj_new}
\end{align}
where the per-head matrices $\vec{O}^{(\ell,m)}$ are defined such that
$\vec{O}^{(\ell)} = [\vec{O}^{(\ell,1)};\ldots;\vec{O}^{(\ell,M)}]$.

It inspires us to propose a parameter-averaging trick that is able to further reduce the storage and computation cost of our method. 
Precisely, \emph{after} training, we average the output projection matrices $\vec{O}^{(\ell,m)}$ as well as the attention parameters inside $\text{Attn}^{(m)}$ across the attention heads. 
Then we only store the averaged parameters. 
During inference, we only use a single attention head into which the stored parameters have been loaded.
That way, although we may have trained $M > 1$ attention heads, our storage and inference cost will be as low as if we had only trained a single head. 
The technical details of this trick is discussed in \cref{app:math}.

\section{Related Work}\label{sec:related}
There are three major paradigms of parameter-efficient transfer learning. 
The first is to only fine-tune a small subset of the existing parameters in a pretrained language model~\citep{howard2018universal,lee2019would,bitfit}. 
The second is adapter-tuning~\citep{houlsby2019parameter}: inserting adapters (i.e., small neural nets) into the language model and only tuning their parameters. 
The third is prefix-tuning~\citep{li-liang-2021-prefix,hambardzumyan-etal-2021-warp,p-tuning,liu2021p,prompt-tuning}: augmenting the input sequence with trainable tokens and only updating the new token embeddings.
Both adapter-tuning and prefix-tuning keeps the PLM frozen. 
Our work falls into the category of adapter-tuning. The key difference is that our proposed adapter has an attention architecture.
The previously proposed methods in this direction all use feed-forward neural networks: they are \citet{houlsby2019parameter,lin-etal-2020-exploring,pfeiffer-etal-2021-adapterfusion,hu2021lora,he2021towards}.

AdaMix~\citep{wang2022adamix} proposes a stochastic routing strategy to mix an ensemble of adapters and it is orthogonal to all the adapter methods including ours. 
Akin to our parameter-averaging trick, they propose to average the adapter parameters for low-cost storage and inference. Similar tricks are used in \citet{ravi2017projectionnet,matena2021merging,wortsman2022model}. 

\section{Experiments}\label{sec:exp}
We evaluated our proposed tiny-attention adapter on a range of natural language understanding tasks including GLUE and FewGLUE. 
Our method is implemented in PyTorch~\citep{pytorch} and heavily relies on HuggingFace~\citep{huggingface}.
Our code is submitted for review and it will be publicly released after the paper is published. 

In all of our experiments, we set the dimensionality of each tiny-attention head (i.e., the dimension of query, key, and value vectors) to be \emph{one}. 
Other experiment details (e.g., hyperparameters) can be found in \cref{app:exp-details}. 

\subsection{Main Results: GLUE and FewGLUE} \label{sec:main}
\paragraph{GLUE.} 
On the GLUE benchmark, we chose the RoBERTa model~\citep{liu2019roberta} as our PLM and we used the pretrained roberta-large weights (355M parameters) downloaded from HuggingFace. 

Our results on the GLUE benchmark are already presented in \cref{fig:para-count}. 
As we can see, our method (Tiny-Attn-1H and Tiny-Attn-4H) outperforms all the previously proposed parameter-efficient tuning methods as well as fine-tuning. 
Yet our method uses significantly fewer trainable parameters than the other methods except WARP. 
The single-head version (Tiny-Attn-1H) trains 176K parameters, which only counts as 0.05\% of the PLM parameters. 
The four-head version (Tiny-Attn-4H) further improves the performance with an increased training cost, but its storage and inference cost remains the same as the single-head version. 

Our method only underperforms AdaMix, which learns a stochastic routing strategy to mix an ensemble of adapters.
But AdaMix uses significantly more trainable parameters than ours. 
Moreover, AdaMix's technique is orthogonal and complementary to most other adapter methods including ours.

Our results on each individual GLUE task can be found in \cref{tab:glue-dev,tab:glue-test} of \cref{app:glue-result-details}.

\paragraph{FewGLUE.}\label{sec:few}
We also evaluated our method on the CB and RTE tasks of the FewGLUE benchmark. 
They are extremely few-shot settings: each task only has 32 training examples.
We chose ALBERT~\citep{albert} as our PLM and we used the pretrained albert-xxlarge-v2 weights (223M parameters) downloaded from HuggingFace.
The detailed setting can be found in \cref{app:glue-exp-details}. 
The result is shown in \cref{tab:few-shot}. It turns out that the performance of our method is comparable to that of PET~\citep{schick-schutze-2021-just} and GPT-3~\citep{gpt3}.
\begin{table}[h!]\centering
\scalebox{0.9}{
\begin{tabular}{lll}\hline
method &CB &RTE \\\hline
Tiny-Attn-1H (ours) &88.57$\pm$2.99 &68.38$\pm$2.53 \\
WARP &87.5 &71.8 \\
PET &85.1 &69.8 \\
iPET &92.9 &74 \\
GPT-3-Small &42.9 &52.3 \\
GPT-3-Medium &58.9 &48.4 \\
GPT-3 &82.1 &72.9 \\
\hline
\end{tabular}}
\caption{Results on the validation set of FewGLUE. We report accuracy for all tasks.}\label{tab:few-shot}
\end{table}

\subsection{Analysis}\label{sec:analysis}
\begin{figure}[!ht]
    \centering
    \includegraphics[width=\columnwidth]{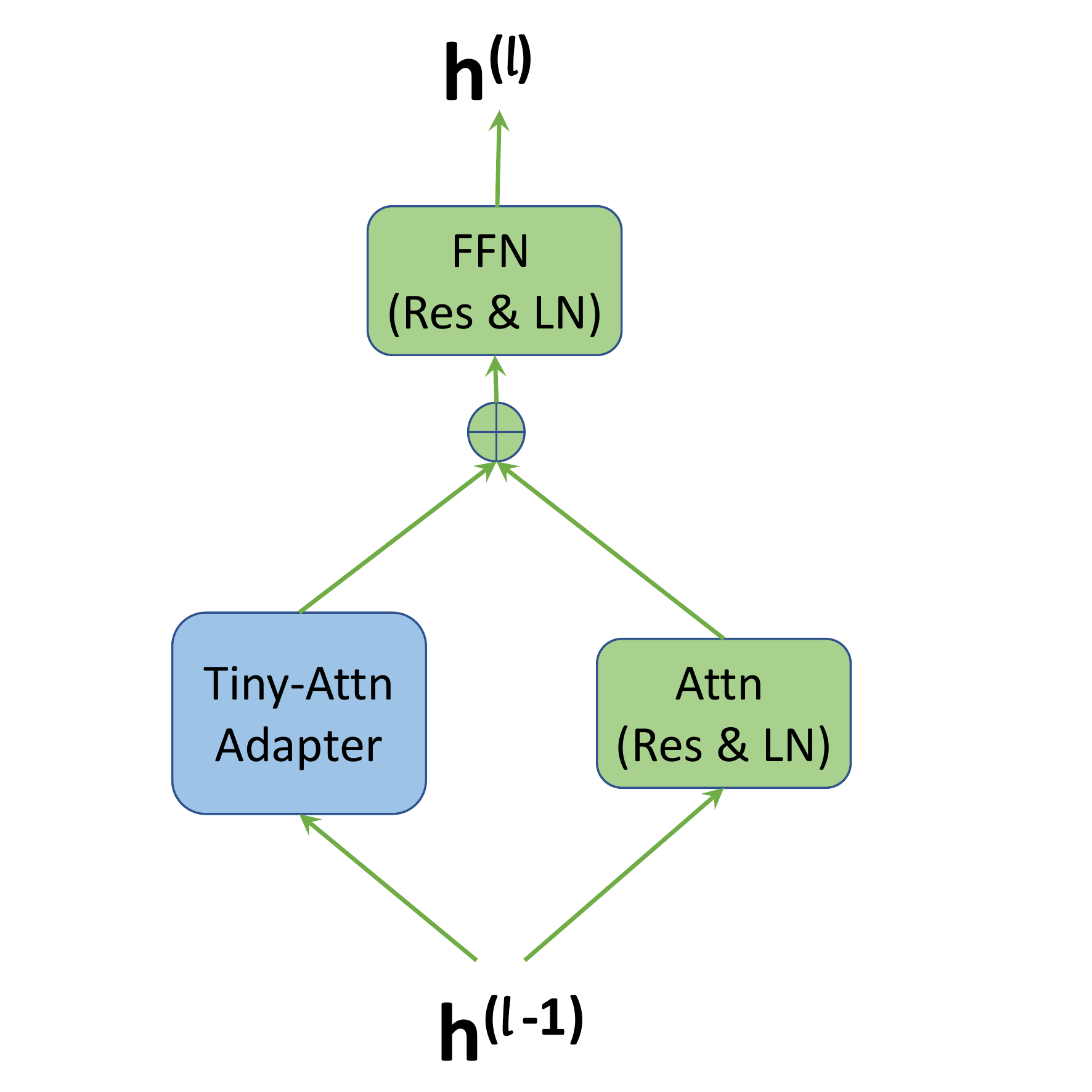}
    \caption{
        Illustration of the parallel structure.
    }
    \label{fig:parallel}
\end{figure}

\paragraph{Sequential vs.\@ parallel.}
In \cref{sec:method}, we presented the `sequential' methods where our tiny-attention modules are placed between the pretrained attention and feed-forward net.
Another option is to put the tiny-attention module in `parallel' to the original attention layer as in \citet{he2021towards}, as illustrated in \cref{fig:parallel}.
We emperically found that these two design choices have negligible differences in results.
Detailed Results are listed in \cref{tab:para} of \cref{app:analysis-result-details}.

\paragraph{Effects of parameter-averaging.}
Recall that we used the parameter-averaging trick in our experiments for an improved inference efficiency. 
But do we lose any performance by using this trick? 
Through ablation studies on CoLA and RTE, we found that using our parameter-averaging trick actually slightly improves the results. 
Detailed results are shown in \cref{tab:weight-averaging} of \cref{app:glue-result-details}.

\begin{figure}[!ht]
    \centering
    \begin{subfigure}[b]{\columnwidth}
        \centering
        \includegraphics[width=\columnwidth]{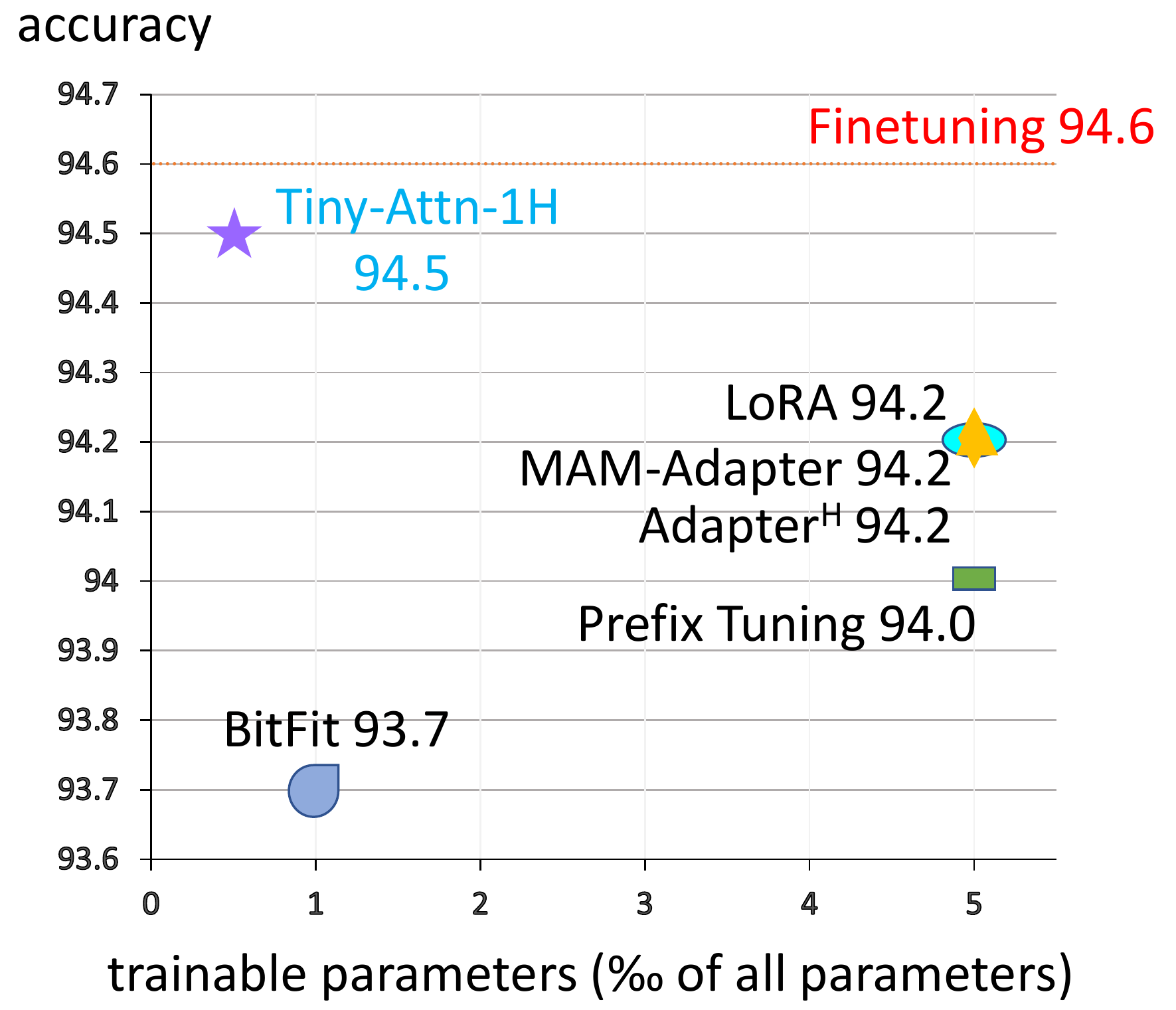}
        \vspace{-15pt}
        \caption{
            SST-2
        }
        \label{fig:sst2}
    \end{subfigure}

    \begin{subfigure}[b]{\columnwidth}
        \centering
        \includegraphics[width=\columnwidth]{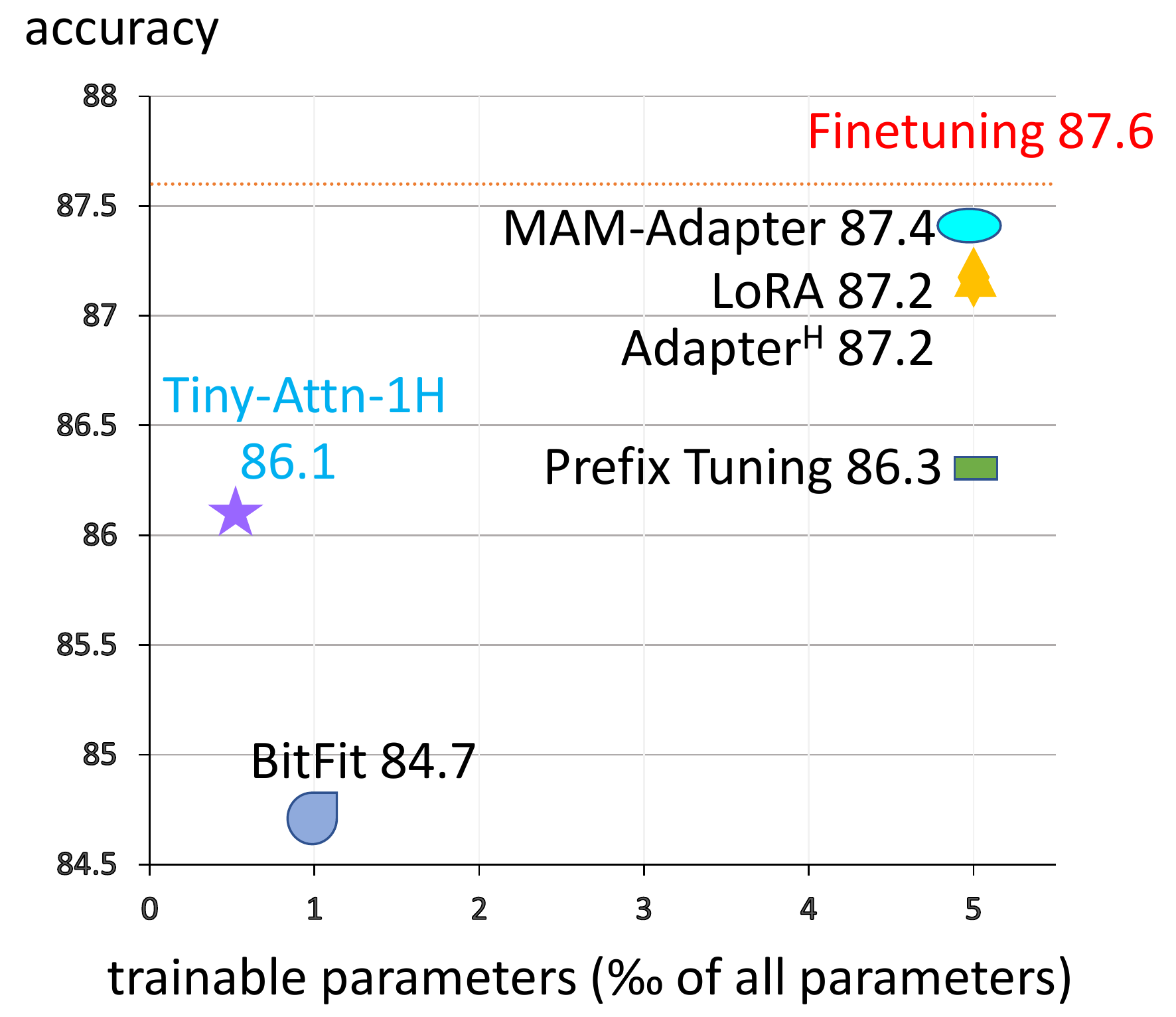}
        \vspace{-15pt}
        \caption{
            MNLI
        }
        \label{fig:mnli}
    \end{subfigure}
    \vspace{-5pt}
    \caption{
        Performance of different parameter-efficient tuning methods on SST-2 and MNLI with RoBERTa-base. The baseline results are from \citet{he2021towards}. 
    }
    \label{fig:size}
\end{figure}

\paragraph{Does the size of PLM matter?}
Parameter-efficient tuning methods are known to suffer performance drop when working with small-sized PLMs. 
To investigate this effect, we also experimented with the pretrained roberta-base (125M parameters) downloaded from Huggingface on the MNLI and SST-2 tasks. 
The results are shown in \cref{fig:size}.
Different methods suffer almost the same amount of performance drop.\footnote{LoRA and Adapter$^{H}$ are the only previous methods that reported GLUE results for both roberta-large and roberta-base.}
\begin{table}[ht]\centering
\small
\begin{tabular}{lccc}\hline
& \multicolumn{3}{c}{performance drop}\\
method & MNLI & SST-2 & average\\\hline
LoRA & 3.4 & 2 & 2.7\\
Adapter$^H$ & 3.1 & 2.1 & 2.6\\
Tiny-Attn-1H (ours) & 3.2 & 2.1 & 2.65\\
\hline
\end{tabular}
\caption{Performance drop of different methods on MNLI and SST-2, with roberta-large and roberta-base as the PLMs.}\label{tab:performance_drop}
\end{table}
But our method enjoys a much larger drop in the number of trainable parameters: the trainable parameters of our method (Tiny-Attn-1H) still only count as 0.05\% of the PLM parameters, but those of LoRA and Adapter$^H$ increase from 0.23\% to 0.5\%. See \cref{tab:performance_drop}.
 
\paragraph{Effects of larger dimensions.}
We set $D=1$ in all our main experiments. It's natural to ask that what if we use a larger dimension. We experimented with a $D=4$ variant of our method, which we call Tiny-Attn-1H4D, on CoLA and RTE. We found that this variant performs slightly worse than the standard version Tiny-Attn-1H. It's normal that more parameters in the adapters could lead to slightly worse performance on some GLUE tasks, see \citet{hu2021lora}.
Detailed results are shown in \cref{tab:1H4D} of \cref{app:glue-result-details}.
 
\section{Conclusion}
In this paper, we presented the tiny-attention adapter. 
While previous adapter-tuning only processes in-place embeddings, our method considers the context from other positions.
Thanks to this contextual modeling, the size of our tiny-attention adapter can be extremely light (e.g., only 1 attention head and 1 head dimension).
To further enable trade-off between performance and training cost, we proposed the weight-averaging technique.
On GLUE benchmark, our tiny-attention adapter achieved better results than full fine-tuning while only updating 0.05\% of parameters. 
Our model also achieved competitive results under the few-shot setting. 
Lastly, we compared our methods with the alternatives (e.g., parallel instead of sequential, without parameter averaging) and also showed the generalization to smaller PLMs.

\section*{Acknowledgements}
This work was supported by a research gift to the last author by Adobe Research. 
We thank the anonymous EMNLP reviewers and meta-reviewer for their constructive feedback. 
We also thank our colleagues at Toyota Technological Institute at Chicago for helpful discussion.

\section*{Limitations}
Our main limitation is that our method was only evaluated on the GLUE and FewGLUE benchmarks and that we haven't experimented with a diverse set of generation tasks (e.g., XSUM~\citep{xsum}, E2E~ \citep{e2e}) yet. 
\citet{he2021towards} reported that the state-of-the-art parameter-efficient adaptation method on a task or dataset may suffer a sharp performance drop on another task or dataset. 
Although our method is consistently effective across multiple classification tasks, it is still possible that it won't perform well at generation tasks such as summarization and translation. 

\section*{Ethics Statement}
Our method belongs to the general category of efficient language model fine-tuning and our focus is to reduce the number of trainable parameters.
It can benefit the scenarios when communications become a bottleneck, such as federal learning, distributed training, and edge computing. 
However, since we do not apply explicit differentiable privacy methods to these updated parameters, the method can be vulnerable to specific attacks (e.g., man-in-the-middle attack). 

Our method can also be deployed to on-device applications, where the storage is limited.
For these applications, different tasks can just keep a small set of task-specific parameters while other parameters are shared. 

Our method can help reduce the carbon footprint of the model training in two ways.
Firstly, the model is fine-tuned from a pretrained model thus can be adapted to a downstream task quickly while reaching a satisfying performance.
Secondly, our method only updates a small portion of all parameters thus the optimizer only tracks a small part of parameters. 
For this reason, the same training infrastructure can support larger batch size given the optimizer's states are significantly reduced (e.g., to less than 1\% in our Tiny-Attn-1H method).

Meanwhile, our method shares the same possibilities as most of previous efficient training methods, such as misusage, containing data bias, and suffering from adversarial attacks. 
However, the method developed in this paper is orthogonal to the previous effort to mitigate the above issues.

\bibliography{tiny_attn}
\bibliographystyle{acl_natbib}

\clearpage
\appendix

\section{Method Details}\label{app:math}
The notations we use in this section are inherited from \cref{sec:method}.

Recall that every head-specific attention vector $\vec{\tilde{z}}_t^{(\ell,m)}$ is given by \cref{eq:attnm}. If we expand $\text{Attn}_t^{(m)}$ in this equation, we get
\begin{align}
\begin{split}
    &\quad\ \text{Attn}_t^{(m)}({\vec{z}}^{(\ell)}_0, {\vec{z}}^{(\ell)}_1, \ldots, {\vec{z}}^{(\ell)}_T) \\
    &= \frac{\sum_{s=0}^T\exp\left(\frac{\vec{q}_t^{(\ell,m)\top}\vec{k}_s^{(\ell,m)}}{\sqrt{D}}\right)\vec{v}_s^{(\ell,m)}}{\sum_{s=0}^T\exp\left(\frac{\vec{q}_t^{(\ell,m)\top}\vec{k}_s^{(\ell,m)}}{\sqrt{D}}\right)}
\end{split}
\end{align}
The subscript $t$ in $\text{Attn}_t^{(m)}$ means that the $t$\th intermediate embedding ${\vec{z}}^{(\ell)}_t$ is used to construct the query $\vec{q}_t^{(\ell,m)}$. Similarly, for $s=0,1,\ldots,T$,  $\vec{k}_s^{(\ell,m)}, \vec{v}_s^{(\ell,m)}$ are low-dimensional key and value vectors that are dependent on the high-dimensional intermediate embedding ${\vec{z}}^{(\ell)}_s$. These vectors are computed by linear transformations:
\begin{subequations}
\begin{align}
    \vec{q}_t^{(\ell,m)} &= \vec{W}_Q^{(\ell,m)}\vec{z}_t^{(\ell)}\\
    \vec{k}_s^{(\ell,m)} &= \vec{W}_K^{(\ell,m)}\vec{z}_s^{(\ell)}\\
    \vec{v}_s^{(\ell,m)} &= \vec{W}_V^{(\ell,m)}\vec{z}_s^{(\ell)}
\end{align}
\end{subequations}
where $\vec{W}_Q^{(\ell,m)},\vec{W}_K^{(\ell,m)},\vec{W}_V^{(\ell,m)}$ are $H \times 1$ parameter matrices of the attention module, where $H$ is the hidden size of the PLM.

\paragraph{Mixture of experts.}  
During inference, we average all the parameters of the attention module across the attention heads.
\begin{subequations}
\begin{align}
    \vec{W}_Q^{(\ell,n)} &\gets \frac{1}{M} \sum_{m=1}^{M} \vec{W}_Q^{(\ell,m)}\\
    \vec{W}_K^{(\ell,n)} &\gets \frac{1}{M} \sum_{m=1}^{M} \vec{W}_K^{(\ell,m)}\\
    \vec{W}_V^{(\ell,n)} &\gets \frac{1}{M} \sum_{m=1}^{M} \vec{W}_V^{(\ell,m)}
\end{align}
\end{subequations}
where $n=1,2,\ldots,M$.

Then the attention function $\text{Attn}_t^{(m)}$ will be the same for all $m$. In other words, we will have $\tilde{\vec{z}}_{t}^{(\ell,1)} = \ldots = \tilde{\vec{z}}_{t}^{(\ell,M)}\defeq \tilde{\vec{z}}_{t}^{(\ell)}$. 

Substitute this to \cref{eq:outproj_new}, we get
\begin{align}
    \tilde{\vec{z}}^{(\ell)}_t 
    &= (\sum_{m=1}^{M} \vec{O}^{(\ell,m)}) \tilde{\vec{z}}^{(\ell)}_t
\end{align}
where $\vec{O}^{(\ell,m)}\in\mathbb{R}^{1 \times H}$ are the output projection matrices. We can define
\begin{align}
    \vec{\bar{O}}^{(\ell)} \defeq \frac{1}{M} \sum_{m=1}^{M} \vec{O}^{(\ell,m)}
\end{align}
Then we only use the averaged projection matrix such that \cref{eq:outproj_new} becomes
\begin{align}
    \vec{z}^{(\ell)}_t 
    &= M \vec{\bar{O}}^{(\ell)} \tilde{\vec{z}}^{(\ell)}_t
\end{align}

\section{Experiment Details}\label{app:exp-details}
We used the PyTorch library~\citep{pytorch} and the pretrained language models from the HuggingFace transformers library~ \citep{huggingface} in all of our experiments.

\subsection{GLUE and FewGLUE}\label{app:glue-exp-details}
\paragraph{GLUE.}
We evaluated our method on 8 tasks of the GLUE benchmark~ \citep{wang2018glue}: SST-2~\citep{sst2}, CoLA~\citep{cola}, MNLI~\citep{mnli}, QQP\footnote{\url{https://quoradata.quora.com/First-Quora-Dataset-Release-Question-Pairs}}, RTE~\citep{rte,rte2,rte3,rte4}, MRPC~\citep{mrpc}, QNLI~\citep{qnli} and STS-B~\citep{stsb}.\footnote{The dataset can be downloaded at \url{https://huggingface.co/datasets/glue}.} They are all classification tasks except STS-B, which is formed as a regression task. WNLI~\citep{wnli} is excluded following prior work~\citep{houlsby2019parameter,hu2021lora}. 

We used the official data splits. The data were pre-processed by GLUE and HuggingFace and we did not apply any extra modification. The sizes of the training sets of each task are shown in \cref{tab:data-size}. All of the tasks have a balanced label distribution in the training set except for MRPC (68\% positive) and QQP (63\% negative). Besides, the test set of QQP has a different label distribution than the training set.

\begin{table}[ht]\centering
\begin{tabular}{cr}\hline
task & |train| \\\hline
SST-2 & 67k\\
CoLA & 8.5k\\
MNLI & 393k\\
QQP & 364k\\
RTE & 2.5k\\
MRPC & 3.7k\\
QNLI & 105k\\
STS-B & 7k\\
\hline
\end{tabular}
\caption{Sizes of training sets of tasks on GLUE. }\label{tab:data-size}
\end{table}

We trained using AdamW~\citep{adamw} and either a linear or a cosine learning rate decay scheduler. We used a linear warmup with approximately 10\% of the total steps as the warmup steps in some tasks. The number of epochs was fixed to 20. We evaluated on the validation set twice per epoch and report the best result. We used the same batch size 16 for all our tasks. For reproducibility, we used a fixed random seed. Learning rate (lr) was selected from [1E-6,1E-5,1E-4,5E-4,8E-4,1E-3,1.5E-3,2E-3,3E-3,5E-3,7E-3] and weight decay (dec) was chosen from [0,0.01,0.02,0.03,0.04,0.05,0.1,0.2,0.3,0.4,0.5]. Note that we did not do a full hyperparameter search. On average, we run about 80 experiments for each task. However, the detailed numbers of experiments are varying according to the size of the dataset to save computational cost. E.g., on MNLI and QQP, we only perform about 20 experiments for each. We performed our experiments mainly on 10 Nvidia RTX A4000. The average running time is about eight hours, varying across tasks.

For Tiny-Attn-1H, we initialized the output projection matrices to be very small ($\mathcal{U}(-\frac{0.01}{\sqrt{D}},\frac{0.01}{\sqrt{D}})$ where D is the per-head dimension) to make our model behave like the original pretrained model at early stages of training. Empirically, we found this trick important in stabilizing the training. The hyperparameters of the best-performing single-head models are shown in \cref{tab:hyperpara1H}. 

\begin{table}[ht]\centering
\begin{tabular}{crlcc}\hline
task &\multicolumn{1}{c}{lr} &dec &warmup &scheduler \\\hline
SST-2 &8E-4 &0.01 &yes &cosine \\
CoLA &5E-4 &0.04 &no &linear \\
MNLI &1E-3 &0 &yes &linear \\
QQP &5E-4 &0.01 &no &linear \\
RTE &2E-3 &0.05 &no &linear \\
MRPC &3E-3 &0.3 &yes &cosine \\
QNLI &8E-4 &0.02 &no &linear \\
STS-B &1.5E-3 &0 &no &cosine \\
\hline
\end{tabular}
\caption{Hyperparameters for Tiny-Attn-1H. }\label{tab:hyperpara1H}
\end{table}

For Tiny-Attn-4H, we perturbed the weight of Tiny-Attn-1H to initialize every head due to computational limits. In principle, we only need to initialize the weight of every head to be almost the same. This makes the parameter-averaging trick applicable since the weight of these heads would not be very far from each other, just as when we average the parameters of multiple fine-tuned models with the same initialization~\citep{same-error-basin}. The hyperparmeters we use are shown in \cref{tab:hyperpara4H}. 

\begin{table}[ht]\centering
\begin{tabular}{crlcc}\hline
task &\multicolumn{1}{c}{lr} &dec &warmup &scheduler \\\hline
CoLA &8E-4 &0.05 &yes &cosine \\
RTE &1E-5 &0.3 &no &linear \\
MRPC &3E-3 &0.3 &no &linear \\
QNLI &1E-6 &0.4 &no &cosine \\
\hline
\end{tabular}
\caption{Hyperparameters for Tiny-Attn-4H. Since we initialize from the weight of Tiny-Attn-1H, tasks on which we don't get an improvement are omitted.}\label{tab:hyperpara4H}
\end{table}

\paragraph{FewGLUE.}
FewGLUE~\citep{schick-schutze-2021-just} is a subset of the SuperGLUE benchmark~\citep{superglue} with the sizes of all training sets being 32.\footnote{The dataset can be downloaded at \url{https://github.com/timoschick/fewglue}.} We evaluated our model on two tasks of it: CB~\citep{cb} and RTE. 
We still used AdamW but we did not use a scheduler. We used a linear warmup with 10\% of the total steps as the warmup steps for RTE. The number of epochs was fixed to 20. Following \citet{schick-schutze-2021-just}, we did not use any additional validation set for parameter selection or early stopping and we only report the final result. In contrast to them, we did not use any additional unlabeled examples. We used a fixed learning rate 1E-3 and a fixed batch size 1. We've tried a few different weight decay [0,0.02,0.05,0.1], but eventually we chose the same value 0.02 for all tasks. After selecting the hyperparameters with a fixed random seed, we ran over 4 additional random seeds and report the average performance and the standard variation.

Following \citet{schick-schutze-2021-just}, we used albert-xxlarge-v2~\citep{albert} downloaded from the HuggingFace library as the backbone PLM, and used manual prompts to replace the classification head. The manual prompt we use was ``[CLS]<hypothesis>?[MASK].<premise><SEP>'', the same as theirs. The language model predicts the token at the masked position and the output space is limited to a few selected tokens. We used ``yes'' to represent ``entailment'', ``no'' to represent ``contradiction'', and ``maybe'' to represent ``neutral''.

\subsection{Analysis} \label{app:analysis-exp-details}
\paragraph{Sequential vs.\@ parallel.}
In the main experiments, we used a `sequential' (seq) structure where our tiny-attention modules are placed between the pretrained attention and feed-forward net.
Another option is to put the tiny-attention module in `parallel' (para) to the original attention layer as in \citet{he2021towards}.

We used the same setting as our main experiments on GLUE except that we used roberta-base as the backbone PLM. Warmup was used for all experiments. The hyperparameters we used are shown in \cref{tab:hyperpara-para}. 
\begin{table}[ht]\centering
\begin{tabular}{ccrlrc}\hline
task& type &\multicolumn{1}{c}{lr} &dec &scheduler \\\hline
SST-2& seq &1E-3 &0.02 &cosine \\
SST-2& para &1E-3 &0.01 &cosine \\
MNLI& seq &1E-4 &0.01 &linear \\
MNLI& para &1E-3 &0.01 &linear \\
\hline
\end{tabular}
\caption{Hyperparameters for experiments with roberta-base as the PLM.}\label{tab:hyperpara-para}
\end{table}

Note that there are other possible placements, e.g., we can place tiny-attention adapters above the feed-forward networks. We did not do this because our philosophy is to augment the pretrained attention but not to intervene anything else. 

\paragraph{Effects of parameter-averaging.}
We used the same setting as our Tiny-Attn-4H on GLUE except that we did not use parameter-averaging during inference. Instead, we used all 4 heads as in training. The hyperparameters we used are shown in \cref{tab:hyperpara-weight-averaging}.
\begin{table}[ht]\centering
\begin{tabular}{lrlrr}\hline
task &lr &dec &warmup &scheduler \\\hline
CoLA &8E-4 &0 &no &cosine \\
RTE &7E-3 &0.5 &yes &cosine \\
\hline
\end{tabular}
\caption{Hyperparameters for Tiny-Attn-4H without parameter-averaging.}\label{tab:hyperpara-weight-averaging}
\end{table}

\paragraph{Does the size of PLM matter?}
The hyperparameters we used are shown in \cref{tab:hyperpara-para}.

\section{Results and Analysis Details}\label{app:result-details}
We use ``Adapter$^H$'' and ``Adapter$^P$'' to denote the Adapter proposed in \citet{houlsby2019parameter} and \citet{pfeiffer-etal-2021-adapterfusion}, respectively. ``Tiny-Attn-kH'' represents our method with k attention heads, e.g., `Tiny-Attn-1H is our method with a single attention head.
\subsection{GLUE and FewGLUE.}\label{app:glue-result-details}
Results in this section are discussed in \cref{sec:main}.

\begin{table*}[tp]\centering
\scalebox{0.68}{
\begin{tabular}{lccd{3/2}d{3/2}cd{3/2}ccd{3/2}rrr}\hline
method &SST-2 &CoLA &\multicolumn{1}{P{1.4cm}}{MNLI} &\multicolumn{1}{P{1.4cm}}{QQP} &RTE &\multicolumn{1}{P{1.4cm}}{MRPC} &QNLI &STS-B &\multicolumn{1}{P{1.4cm}}{average} &\# trainable para& percentage\\
\hline
metric &Acc &Mat & \text{All}/\text{M} & \text{Acc}/\text{F1} & Acc & \text{Acc}/\text{F1} & Acc & Pearson & \  &\ &\ \\
\hline
AdaMix & 97.1 & 70.2 & 90.9/- & 92.3/89.8 & 89.2 & 91.9/94.1 & 95.4 & 92.4 & 89.9/- & 2M & 0.56\%\\
\hline
Adapter$^H$ & 96.3 & 66.3 & 90.3/- & 91.5/- &72.9 & 87.7/- & 94.7& 91.5 &86.4/- & 0.8M &0.23\%\\
Adapter$^H$ & 96.2 & 66.5 & 89.9/- & 92.1/- &83.4 & 88.7/- & 94.7& 91.0 &87.8/- & 6M &1.7\%\\
Adapter$^P$ & \textbf{96.6} & 67.8 & 90.5/- & 91.7/- & 80.1 & 89.7/- & 94.8 & 91.9 & 87.9/- & 0.8M &0.23\%\\
Adapter$^P$ & 96.1 & 68.3 & 90.2/- & 91.9/- & 83.8 &90.2/- & 94.8 & 92.1 & 88.4/- & 3M &0.85\%\\
LoRA & 96.2 & 68.2& \textbf{90.6}/- & 91.6/- &85.2& 90.2/- & \textbf{94.8} &\textbf{92.6} &88.6/- &0.8M &0.23\%\\
WARP &96.0 &60.6 &-/88.2 &-/84.5 &75.8 &-/90.8 &93.5 &88.6 &-/84.75 &25K &0.007\%\\
fine-tuning &96.4 &68.0 &90.2/90.2 &\textbf{92.2}/- &86.6 &90.9/- &94.7 &92.4 &88.9/- &355M &100\%\\
Tiny-Attn-1H (ours) &\textbf{96.6} &68.8 &89.3/89.6 &90.1/86.6 &88.8 &92.4/94.4 &94.3 &92.3 &89.1/88.9 &176K &0.05\%\\
Tiny-Attn-4H (ours) &\textbf{96.6} &\textbf{69.4} &89.3/89.6 &90.1/86.6 &\textbf{89.2} &\textbf{92.6}/\textbf{94.6} &94.5 &92.3 &\textbf{89.3}/\textbf{89.1} &0.7M &0.2\%\\
\hline
\end{tabular}}
\caption{Dev set results on GLUE tasks. All the runs use roberta-large as the backbone. We report Matthew’s correlation for CoLA, the overall(matched and mismatched) accuracy and matched accuracy for MNLI, Pearson correlation for STS-B, accuracy and F1 score for QQP and MRPC, and accuracy for other tasks. Higher is better for all metrics. All the runs follow the setting in \citet{houlsby2019parameter} except fine-tuning. The results of WARP and AdaMix are taken from their own paper respectively. Other results are taken from \citet{hu2021lora}. }\label{tab:glue-dev}
\end{table*}

\begin{table*}[tp]\centering
\scalebox{0.65}{
\begin{tabular}{lccd{3/2}d{3/2}cd{3/2}cccrl}\hline
method &SST-2 &CoLA &\multicolumn{1}{P{1.3cm}}{MNLI} &\multicolumn{1}{P{1.3cm}}{QQP} &RTE &\multicolumn{1}{P{1.3cm}}{MRPC} &QNLI &STS-B &score &\# trainable para & backbone model\\\hline
metric &Acc &Mat & \text{M}/\text{MM} & \text{Acc}/\text{F1} & Acc & \text{Acc}/\text{F1} & Acc & \text{Pearson}/\text{Spearmanr} & \  &\ &\ \\
\hline
human baselines &97.8 & 66.4 & 92.0/92.8 & 59.5/80.4 & 93.6 & 86.3/80.8 & 91.2 & 92.7/92.6 &87.1 & - & -\\
fine-tuning&97.5 &71.5 & 91.9/91.6 & 76.2/90.8 & 93.2 & 94.0/92.0 & 99.2 & 92.9/92.6 &90.8 & 1.5B& DeBERTa\\
fine-tuning&96.7 & 67.8 & 90.8/90.2 & 74.3/90.2 & 88.2 & 92.3/89.8 & 95.4 & 92.2/91.9 & 88.1 & 355M&roberta-large\\
fine-tuning& 94.9 & 60.5 & 86.7/85.9 & 72.1/89.3 & 70.1 & 89.3/85.4 & 92.7 & 87.6/86.5 & 80.5 & 345M&bert-large\\
WARP &96.3 &53.9 &88.0/88.2 &68.6/87.7 &84.3 &88.2/83.9 &93.5 &89.5/88.8 &81.6 &25K& roberta-large\\
Tiny-Attn-1H (ours) &96.2 &62.5 &89.3/88.8 &71.8/89.1 &83.6 &90.7/87.5 &94.4 &91.1/90.3 &83.5 & 176K & roberta-large\\
\hline
\end{tabular}}
\caption{Test set results on GLUE tasks. We use the same setting as on the dev set, different from the fine-tuning baseline and WARP. The results other than ours are published in \citet{hambardzumyan-etal-2021-warp}. We don't show the result of WNLI but it's considered in the final score.
}\label{tab:glue-test}
\end{table*}
\paragraph{GLUE.}
\cref{tab:glue-dev} shows our results on GLUE development set with roberta-large as the PLM. Note that WARP used a slightly different set of validation metrics from other methods, but based on our results we can assume that this difference does not make a significant difference in the average score, that's why we directly used its score in \cref{fig:para-count}. We count all the parameters updated in training for AdaMix and our Tiny-Attn-4H to compare with other methods. 

We report Matthew’s correlation for CoLA, the overall(matched and mismatched) accuracy and matched accuracy for MNLI, Pearson correlation for STS-B, accuracy and F1 score for QQP and MRPC, and accuracy for other tasks. \footnote{The implementation of these metrics can be found at \url{https://github.com/huggingface/datasets/blob/master/metrics/glue/glue.py}.} We used matched accuracy for MNLI and accuracy for QQP and MRPC as the validation metrics. For other tasks, we just used the metrics we report.

To make a fair comparison, we did not use a model pretrained on MNLI for MRPC, RTE, and STS-B as the fine-tuning baseline did, following the setting in \citet{houlsby2019parameter} and \citet{hu2021lora}. Under this restriction, our method still outperforms the fine-tuning baseline on average. 

There are some other parameter-efficient transfer learning methods evaluated on GLUE, but they either use a different backbone, e.g. \citet{karimi2021compacter}; or they are orthogonal to our work, e.g. \citet{ruckle-etal-2021-adapterdrop}, so we do not include them in this table.

We obtained test set results on the official GLUE server. We only made one submission to the server. Following WARP, we always predict the majority class for WNLI. For other tasks, we directly used the weight with the highest score on the validation sets and did not do any extra modification as the fine-tuning baseline did. Specially, we did not use the model adapted to MNLI for RTE, MRPC and SST-B as WARP and the fine-tuning baseline did. We found that our method has a consistent performance on validation and test sets, comparing to WARP. The results are shown in \cref{tab:glue-test}. Note that the score in the table is the official GLUE score, so it is affected by the result of WNLI.

\subsection{Analysis}\label{app:analysis-result-details}
The results in this section are discussed in \cref{sec:analysis}. 

\paragraph{Sequential vs.\@ parallel.}
\cref{tab:para} shows our results on SST-2 and MNLI with different structures. We found that these two design choices have negligible differences in results.
\begin{table}[h!]\centering
\begin{tabular}{cll}\hline
method &SST-2 &MNLI \\\hline
sequential &94.5 &86.1 \\
parallel &94.4 &86.1 \\
\hline
\end{tabular}
\caption{Performance of Tiny-Attn-1H with different structures on SST-2 and MNLI. We report accuracy for SST-2 and matched accuracy for MNLI.}
\label{tab:para}
\end{table}

\paragraph{Effects of parameter-averaging.}
We found that using our parameter-averaging trick actually slightly improved the results, as shown in \cref{tab:weight-averaging}. 
\begin{table}[h!]\centering
\begin{tabular}{cll}\hline
parameter-averaging &CoLA &RTE \\\hline
no &68.8 & 88.8\\
yes &69.4 & 89.2\\
\hline
\end{tabular}
\caption{Results of Tiny-Attn-4H with different parameter-averaging settings on CoLA and RTE. We report Matthew’s correlation for CoLA and accuracy for RTE.}
\label{tab:weight-averaging}
\end{table}

\paragraph{Effects of larger dimensions}
We found that Tiny-Attn-1H4D is slightly worse than the standard version Tiny-Attn-1H, as shown in \cref{tab:1H4D}. 
\begin{table}[h!]\centering
\begin{tabular}{cll}\hline
method &CoLA &RTE \\\hline
Tiny-Attn-1H4D &68.0 & 87.4\\
Tiny-Attn-1H &68.8 & 88.8\\
\hline
\end{tabular}
\caption{Results of Tiny-Attn-4H with different parameter-averaging settings on CoLA and RTE. We report Matthew’s correlation for CoLA and accuracy for RTE.}
\label{tab:1H4D}
\end{table}

\paragraph{Stability test}
We report the best result with a fixed random seed for our main experiments. To test the stability of our method, we run experiments on SST-2 and MNLI using the same hyperparameters with 5 different random seeds. The results are shown in \cref{tab:stab-test}. We could see that the results on larger datasets like MNLI are quite stable, while the results on SST-2 have a larger variance.
\begin{table}[h!]\centering
\begin{tabular}{cll}\hline
 &SST-2 &MNLI \\\hline
seed-1 &96.6 & 89.6\\
seed-2 &96.6 & 89.5\\
seed-3 &95.8 & 89.6\\
seed-4 &96.3 & 89.8\\
seed-5 &95.2 & 89.6\\\hline
avg &96.1 & 89.6\\
stdev &0.6 &0.1\\
\hline
\end{tabular}
\caption{Stability test. We report accuracy for SST-2 and matched accuracy for MNLI. Seed-1 is the seed we used in the main experiments.}
\label{tab:stab-test}
\end{table}

\paragraph{Generation results}
There has been research showing that parameter-efficient transfer methods with good classification performance may not work equally well for non-classification tasks, and vice versa~\citep{he2021towards}. We conducted experiments on E2E NLG Challenge~\citep{novikova2017e2e} and found that our method also suffers from this problem. The results are shown in \cref{tab:E2E}. Our result is comparable to adapter but worse than prefix-tuning.
\begin{table}[h!]\centering
\begin{tabular}{cllr}\hline
method &dev &test &\# trainable para\\\hline
Tiny-Attn &71.4 & 63.8& 0.1M\\
Adapter &68.1 & 66.3& 0.36M\\
Prefix-Tuning &74.8 &70.3& 0.36M\\
\hline
\end{tabular}
\caption{Results on E2E NLG Challenge benchmark with gpt2-medium as the PLM. We report the BLEU score computed by the official evaluation script.}
\label{tab:E2E}
\end{table}

\end{document}